\documentclass[letterpaper, 10pt, conference]{ieeeconf}

\IEEEoverridecommandlockouts                              %

\usepackage{graphics} %
\usepackage{epsfig} %
\usepackage{mathptmx} %
\usepackage{times} %
\usepackage{amsmath} %
\usepackage{amssymb}  %
\usepackage{soul}
\usepackage{xcolor}
\usepackage{dsfont}
\usepackage{hyperref}
\usepackage[printonlyused]{acronym}
\usepackage{mathtools}

\DeclarePairedDelimiter\floor{\lfloor}{\rfloor}
\usepackage{bm}
\DeclareMathOperator*{\argmin}{arg\,min}
\title{\LARGE \bf
Embedded IPC: Fast and Intersection-free Simulation in Reduced Subspace for Robot Manipulation}

\author{Wenxin Du$^{*1}$, Chang Yu$^{*1}$, Siyu Ma$^{1,4}$, Ying Jiang$^{1}$, Zeshun Zong$^{1}$, \\Yin Yang$^{3}$, Joe Masterjohn$^{2}$, Alejandro Castro$^{2}$, Xuchen Han$^{2}$, Chenfanfu Jiang$^{1}$
\thanks{* equal contribution.}
\thanks{$^{1}${\tt\footnotesize \{wenxindu,changyu1,yingjiang,zeshunzong,cffjiang\} @ucla.edu}, AIVC Laboratory, UCLA, USA.}
\thanks{$^{2}${\tt\footnotesize \{joe.masterjohn,alejandro.castro,xuchen.han\} @tri.global}, Toyota Research Institute, USA.}
\thanks{$^{3}$ {\tt\footnotesize yin.yang@utah.edu}, University of Utah, USA.}
\thanks{$^{4}$ {\tt\footnotesize sim003@ucsd.edu}, UCSD, USA.}
}

\begin{document}

\newcommand{\vf}[1]{{\bm{#1}}}
\newcommand{\mf}[1]{{\mathbf{#1}}}

\acrodef{dof}[DoF]{Degrees of Freedom}
\acrodef{fem}[FEM]{Finite Element Method}
\acrodef{ipc}[IPC]{Incremental Potential Contact}
\acrodef{abd}[ABD]{Affine Body Dynamics}
\acrodef{ccd}[CCD]{Continuous Collision Detection}
\acrodef{toi}[TOI]{Time of Impact}
\acrodef{nerf}[NeRF]{Neural Radiance Fields}
\acrodef{gs}[GS]{Gaussian Splatting}
\acrodef{vae}[VAE]{Variational Encoder}
\acrodef{accd}[ACCD]{Accumulated Continuous Collision Detection}

\maketitle
\thispagestyle{empty}
\pagestyle{empty}

\begin{abstract}
Physics-based simulation is essential for developing and evaluating robot manipulation policies, particularly in scenarios involving deformable objects and complex contact interactions. However, existing simulators often struggle to balance computational efficiency with numerical accuracy, especially when modeling deformable materials with frictional contact constraints. We introduce an efficient subspace representation for the Incremental Potential Contact (IPC) method, leveraging model reduction to decrease the number of degrees of freedom. Our approach decouples simulation complexity from the resolution of the input model by representing elasticity in a low-resolution subspace while maintaining collision constraints on an embedded high-resolution surface. Our barrier formulation ensures intersection-free trajectories and configurations regardless of material stiffness, time step size, or contact severity. We validate our simulator through quantitative experiments with a soft bubble gripper grasping and qualitative demonstrations of placing a plate on a dish rack. The results demonstrate our simulator's efficiency, physical accuracy, computational stability, and robust handling of frictional contact, making it well-suited for generating demonstration data and evaluating downstream robot training applications. More details and supplementary material are on the website: \url{https://sites.google.com/view/embedded-ipc}. 
\end{abstract}

\section{INTRODUCTION}

Physics-based simulation plays a pivotal role in bridging the gap between real-world and virtual environments, making it an essential tool for learning and evaluating robotic manipulation policies. By offering a safe and controlled virtual space, these simulations allow robots to interact with everyday objects and industrial production environments in a low-cost, risk-free, and efficient manner. Moreover, physics-based simulations enable the large-scale generation of demonstration data for downstream applications. This data can be collected through teleoperation within virtual environments or through automated generation, with simulator-based filtering enhancing the quality and relevance of the data.

\begin{figure} [t]
\centering
\includegraphics[width=0.8\linewidth]{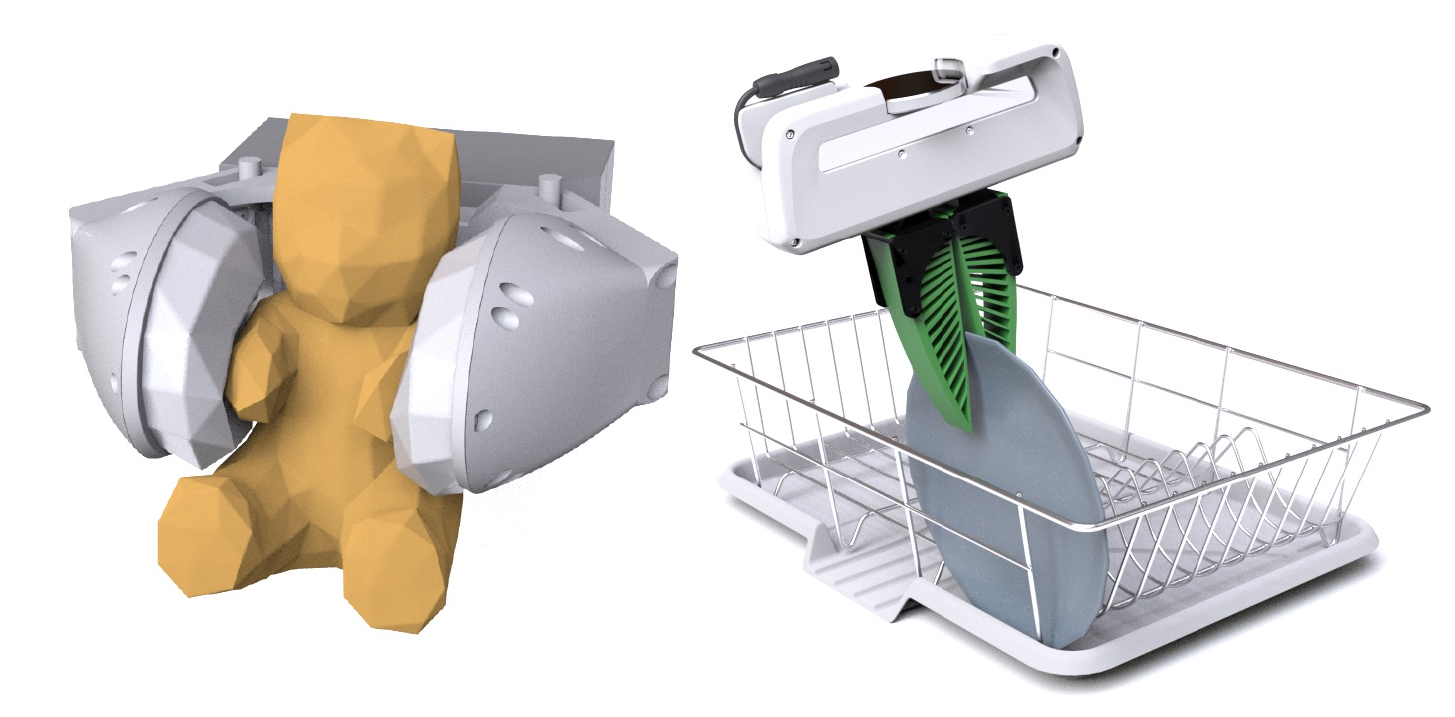}
\vspace{-5mm}
  \caption{Our method can simulate grasping a deformable teddy bear with a bubble gripper and manipulating stiff plates with a FinRay gripper, all while ensuring a non-penetration guarantee at interactive rates.}
  \vspace{-5mm}
  \label{fig:teaser}
\end{figure}

The advancement of soft robotic systems~\cite{schmitt2018soft}, deformable object manipulation~\cite{zhu2022challenges}, and vision-based tactile sensors~\cite{agarwal2021simulation,ding2021sim} opens up new possibilities for developing manipulation policies that can be effectively transferred from simulation to the real world~\cite{kadian2020sim2real}. Although several simulation tools have been developed~\cite{todorov2012mujoco, makoviychuk2021isaac, castro2022unconstrained, han2023convex, faure2012sofa, duriez2013control}, significant challenges remain. Most existing methods and software primarily model robotic grippers and objects using rigid body assumptions and dynamics, which limits their applicability to more complex scenarios. An ideal simulator for object manipulation should meet three critical accuracy requirements: it must integrate unified soft and rigid body dynamics, provide intersection-free guarantees, and accurately model frictional contacts. From a practical standpoint, the simulator’s runtime must be fast enough to facilitate efficient data generation~\cite{wang2023dexgraspnet} for deep learning and reinforcement learning tasks~\cite{zhao2020sim}. However, current simulators either run too slowly, experience penetration issues in contact-rich environments, or lack support for continuum mechanics-based elasticity simulations with precise frictional contact modeling.

\ac{ipc}~\cite{li2020incremental} has proven to be highly effective in simulating elastic materials with a guarantee of intersection-free configurations \cite{kim2022ipc, du2024intersection, du2024tacipc, chen2024general, fernandez2024stark}. However, the original full-space \ac{ipc} method faces performance challenges, especially when simulating deformable objects with large of \ac{dof}s in the system. While modeling a deformable body as a rigid body simplifies the problem, it introduces unrealistic assumptions. On the other hand, assigning \ac{dof} to each vertex in a mesh significantly increases computational complexity. Many robotic manipulation tasks, particularly those involving tools like grippers~\cite{chi2024universal} or spatulas, deal with stiff materials that exhibit relatively simple deformation behaviors. Furthermore, while reducing the degree of deformability can enhance performance, maintaining a detailed mesh topology is still crucial for accurate collision detection and handling.

In this paper, we aim to enhance the efficiency of barrier-based optimization methods by leveraging the strengths of subspace reduction. Our approach introduces an efficient subspace representation for \ac{ipc}, streamlining computation by focusing on a reduced set of coordinates, thereby decoupling simulation complexity from the input model’s resolution without sacrificing contact handling quality. We describe the deformable model using a reduced subspace of tetrahedra with a small number of \ac{dof}s, significantly reducing
the overall computational cost. Notably, collision-related constraints are defined on the surface triangles and vertices of the original body, preserving the non-penetration guarantee as in the original full-space \ac{ipc} method. In summary, our contributions are: 

\begin{itemize}
    \item  \emph{Embedded IPC}, the first strictly non-intersection guaranteeing rigid-deformable robotics simulator that achieves interactive rates. Our approach introduces an efficient subspace representation that leverages subspace reduction, significantly lowering DoFs and decoupling simulation complexity from the resolution of input model.
    \item We validate our simulator through quantitative experiments involving grasping with a soft bubble gripper and qualitative experiments on plate placement in a bowl rack, demonstrating its efficiency, physical accuracy, computational stability, and adherence to the non-penetration guarantee.
\end{itemize}

\section{RELATED WORK}

\subsection{Contact Modeling}
A common Lagrangian approach for discretizing the governing equations for elastic deformable body simulation is the \ac{fem} method~\cite{bathe2007finite}, but an inherent challenge lies in accurately resolving frictional contacts due to the non-smooth nature of the solutions to inelastic constraints. The pyramid approximation of the Coulomb friction cone~\cite{otaduy2009implicit} is often applied to incorporate friction into constrained optimization formulations, but it leads to nonphysical anisotropy of friction. This approximation is introduced to write a linear complementarity programming (LCP)~\cite{baraff1994fast, stewart2000rigid, kaufman2008staggered, daviet2011hybrid}. While work on the existence and uniqueness of LCP solutions is vast, in practice, very often the required mathematical conditions are not satisfied or the LCP is badly ill conditioned. Moreover, the non-existence of solutions~\cite{baraff1993issues}, exponential worst-case complexity, and NP-hardness have led researchers to seek alternative formulations. Using penalty impulses is a popular alternative for preventing penetrations~\cite{cundall1979discrete, terzopoulos1987elastically, bridson2002robust, harmon2008robust}. Rather using inequality constraints, penalty methods allow for slight interpenetration but penalize them using spring-like repulsive forces. These methods involve tuning parameters that lack direct physical interpretation, making it quite difficult in practice to tune to different applications. To make the problem mathematically tractable, \cite{castro2022unconstrained} introduces a convex approximation of a physical model of compliant contact. The method is extended to model continuous compliant surfaces in \cite{masterjohn2022velocity} and to model deformable objects in \cite{han2023convex}. One of the limitations of compliant contact however, is its inability to model thin or even co-dimensional models. Constraint-based methods, such as Position-Based Dynamics~\cite{muller2007position, macklin2016xpbd} and Projective Dynamics~\cite{bouaziz2014projective}, are favored for real-time applications due to their interactive capabilities. Slow first-order convergence and inaccurate material modeling are typical drawbacks of constraint-based methods and often lead to severe artifacts when materials are stiff. They also suffer from penetration issues in contact-rich scenes. Recently, \ac{ipc}~\cite{li2020incremental} introduced barrier functions to provide a penetration free solution and was later extended to rigid-body dynamics~\cite{ferguson2021intersection, lan2022affine}. While \ac{ipc} ensures penetration free solutions, it does so by allowing \emph{action at a distance}, with contact forces that activate when bodies are within a certain distance threshold. Within that threshold, the method is compliant~\cite{drake_fric_0}. Still, IPC was proven very effective in a variety of multibody simulations, including thin-objects.

\subsection{Model Reduction}
Model reduction, or reduced-order modeling, reduces computational cost by projecting high-dimensional \ac{dof}s onto a low-dimensional subspace. A wise choice of the subspace is critical to both the performance and the quality of model reduction. Classic model reduction methods focused on linear methods, such as principal component analysis~\cite{berkooz1993proper}, proper orthogonal decomposition~\cite{barbivc2008real}, and modal analysis~\cite{pentland1989good, hauser2003interactive}. Relatedly, \cite{lee2020model, chen2022crom, zong2023neural} explored nonlinear low-dimensional manifolds, primarily via neural networks. \cite{capell2002interactive} coarsens the simulated geometries to efficiently prescribe the dynamics of skin rigging. In-simulation adaptive retrieving~\cite{narain2012adaptive, li2018implicit} can also reduce unimportant \ac{dof}s while maintaining reasonable accuracy. For the \ac{ipc} family, \cite{lan2021medial} built subspace based on the medial axis transform and \cite{lan2022affine} can be viewed as single-affine-body reduction.

\section{METHOD}

\subsection{Subspace Simulation}

Given a discretization of $N_v$ vertices with positions $\vf{x}_1, \vf{x}_2, ..., \vf{x}_{N_v}$ in the Cartesian space, the solution to the problem of interest in the full space (Cartesian space) is denoted by the stacked position vector $\mf{x} = [\vf{x}^T_1, \vf{x}^T_2, ..., \vf{x}^T_{N_v}]^T.$ We construct a low-dimensional subspace $Q \subset \mathbb{R}^{N_s}$ where $N_s \ll 3N_v,$ and an associated embedding map  $\phi: Q \rightarrow \mathbb{R}^{3N_v}$ that maps from the subspace to the full space as $\mf{x} = \phi(\mf{q})$ for $\mf{q} \in Q.$ The choice of $Q$ and $\phi$ in our method will be deferred to Sec. \ref{sec:embedded_ipc}. 
Solving the dynamics in a subspace introduces constraints into the system. To address this, we use Lagrangian mechanics. The Lagrangian of the reduced system is $L(\mf{q}, \dot{\mf{q}}) = T(\mf{q}, \dot{\mf{q}}) - V(\mf{q})$, where $T(\mf{q}, \dot{\mf{q}})$ and $V(\mf{q})$ are the kinematic energy and potential energy of the system, respectively. The kinetic energy can be written as 
\begin{equation}
    \begin{split}
        T(\mf{q}, \dot{\mf{q}}) &= \frac{1}{2} \mf{\dot{x}}^T \mf{M} \mf{\dot{x}} = \frac{1}{2} \dot{\phi}(\mf{q})^T \mf{M}\dot{\phi}(\mf{q}) =  \frac{1}{2} (\mf{J\dot{q}})^T \mf{M} (\mf{J\dot{q}})\\& = \frac{1}{2}\mf{\dot{q}}^T(\mf{J}^T\mf{M}\mf{J})\mf{\dot{q}}= \frac{1}{2}\mf{\dot{q}}^T \mf{M^q} \mf{\dot{q}},
    \end{split}
\end{equation}
where $\mf{J} = \frac{\partial \phi}{\partial \mf{q}}\in \mathbb{R}^{3N_v \times N_s }$ is the Jacobian matrix, $\mf{M}$ is the full-space mass matrix, and $\mf{M^q} = \mf{J}^T\mf{M}\mf{J}$ is the mass matrix in the subspace. Potential energy $V(\mf{q})$ includes an elastic energy term $\Phi^{\mf{q}}(\mf{q})=\Phi^{\mf{x}}(\phi(\mf{x}))$ and an external force (e.g., gravity) term $E_{\text{ext}}(\mf{q})$. Here $\Phi^{\mf{x}}(\mf{x})=\int_{\Omega} \Psi(\mf{x})d\mf{x}$. $\Psi(\mf{x})$ is the elastic energy density. $\Omega$ is the volume region of all objects in the rest configuration. 

Substituting $L(\mf{q}, \mf{\dot{q}})$ into the Euler-Lagrange equation $\frac{\partial L}{\partial \mf{q}}(\mf{q}, \mf{\dot{q}}) - \frac{d}{dt}\frac{\partial L}{\partial \mf{q}}(\mf{q}, \mf{\dot{q}}) = 0$ yields
\begin{equation}
\mf{M^q}\mf{\ddot{q}} = -\frac{dV}{d\mf{q}}(\mf{q}).
\label{eqn:euler-lagrangian}
\end{equation}
We temporally discretize Eq. (\ref{eqn:euler-lagrangian}) by backward Euler as 
\begin{equation}
    \frac{\mf{q}^{n+1} - \mf{q}^{n}}{h} = \mf{\dot{q}}^{n + 1}, \quad \frac{\mf{M^q}(\mf{\dot{q}}^{n+1} - \mf{\dot{q}}^{n})}{h} = -\frac{dV}{d\mf{q}}(\mf{q}^{n+1}),
\end{equation}
where time is discretized into a sequence of time steps $\{t_n = nh: n\in \mathbb{N}\}$ with time step size $h > 0,$ and $\mf{q}^{n}=\mf{q}(t_n)$. Under this discretization, Eq. \eqref{eqn:euler-lagrangian} can be formulated as 
\begin{equation}
\frac{d}{d \mf{q}}\left(E_{\text{IP}}(\mf{q}^n)\right) = 0
\end{equation}
if we define 
\begin{equation}
    E_{\text{IP}}(\mf{q}) = \frac{1}{2}(\mf{q} - \mf{q}^{n} - h\mf{\dot{q}}^{n})^T\mf{M^q}(\mf{q} - \mf{q}^{n} - h\mf{\dot{q}}^{n}) + h^2V(\mf{q})
\end{equation}
to be the incremental potential energy of the constrained system. The general subspace simulation problem in a conservative system can be reformulated as the minimization problem
\begin{equation}
\mf{q}^{n+1} = \argmin_{\mf{q}} E_{\text{IP}}(\mf{q}).
\end{equation}

\subsection{Frictional Contact}
We adopt the \ac{ipc} for contact handling \cite{li2020incremental}.
Let $\mathcal{B}$ denote all surface point-triangle pairs and edge-edge pairs in object surface meshes. In full-space \ac{ipc}, given a configuration $\mf{x}$, for each point-triangle or edge-edge contact pair $k\in \mathcal{B}$ with distance $d_k>0$, the barrier energy $b(d_k(\mf{x}))$ is 
\begin{equation}
    b(d_k(\mf{x}))=-\left(d_k - \hat{d}\right)^2\log(\frac{d_k}{\hat{d}})I_{\{d_k \in (0, \hat{d})\}}(d_k)
    \label{eqn:IPC_energy}
\end{equation}
and the approximated friction potential energy $D_k(\mf{x}, \mf{x}^{n})$ is 
\begin{equation}
D_k(\mf{x}, \mf{x}^{n}) = \mu\lambda_k^{n}f_0(\lVert \vf{u}_k\rVert),
\end{equation}
where $\mf{x}^{n}$ is the configuration at the last time step $t_n$. 
Here $\hat{d}>0$ is a threshold distance at which IPC contact force application begins; $I(\cdot)$ is the indicator function; $\lambda_k^{n}$ is the magnitude of lagged normal contact force at the previous timestep;
$\vf{u}_k\in \mathbb{R}^2$ is the tangential relative displacement vector in a local orthogonal frame for the contact pair $k$; $f_0(x)=\int_{\epsilon_vh}^x f_1(y)dy + \epsilon_{v}h$ is an integrable approximation of the dynamic-static friction transition, with $f_1(y)$ is given by:
\begin{equation}
f_1(y) = 
\begin{cases}
-\frac{y^2}{\epsilon_v^2h^2} + \frac{2y}{\epsilon_vh}, &y\in(0, h\epsilon_v),\\
1, & y\geq h\epsilon_v.
\end{cases}
\end{equation}
Here $\epsilon_v > 0$ is a velocity magnitude threshold. Any contacts with relative velocities below $\epsilon_v$ are treated as static frictional contacts. 
We refer to \cite{li2020incremental} for more details on the algorithm and derivation of IPC's barrier and friction energies.

As in full-space \ac{ipc}, we can add these two terms into our incremental potential energy for the subspace system: 
\begin{align}
E_{\text{IPC}}(\mf{q}) &= E_{\text{IP}}(\mf{q}) + h^2B(\mf{x}) + h^2D(\mf{x}, \mf{x}^{n}) \\ 
&= E_{\text{IP}}(\mf{q}) + h^2B(\phi(\mf{q})) + h^2D(\phi(\mf{q}), \mf{x}^{n}).
\end{align}
where $B(\vf{x})=\kappa\sum_{k\in \mathcal{B}} b(d_k(\vf{x}))$, $D(\vf{x}, \vf{x}^n)=\sum_{k\in\mathcal{B}} D_k(\vf{x}, \vf{x}^n)$, and $\kappa > 0$ is a stiffness parameter for contacts. 
Optimizing this barrier-augmented incremental potential contact energy gives the simulation results at next time step: 
\begin{equation}
\mf{q}^{n+1} = \argmin_{\mf{q}}~E_{\text{IPC}}(\mf{q}).
\end{equation}

\subsection{Embedded \ac{ipc}}
\label{sec:embedded_ipc}
We have developed a reduced subspace framework of the \ac{ipc} simulations. Now, we will derive a concrete algorithm within this framework.

As in full-space \ac{ipc}~\cite{li2020incremental}, we utilize the Projected Newton method to optimize the embedded \ac{ipc} energy in Eq. \ref{eqn:IPC_energy}. To ensure intersection-free guarantee and maintain configurations within the feasible region required for the interior point method, we apply the \ac{ccd} algorithm during each Newton step to detect all contact pairs that may potentially cause intersections and clamp the step size for line search. Our chosen embedding map $\mf{x}=\phi(\mf{q})=\mf{J}\mf{q}$ is linear with regard to $\mf{q}$, allowing us to apply a highly-efficient \ac{accd}~\cite{tang2009c} algorithm. The detailed construction of $\phi(\mf{q})$ is elaborated in Sec. \ref{sec:embedding_impl}.

\subsection{Embedding Implementation}
\label{sec:embedding_impl}
\begin{figure}
\centering
\includegraphics[width=0.5\linewidth]{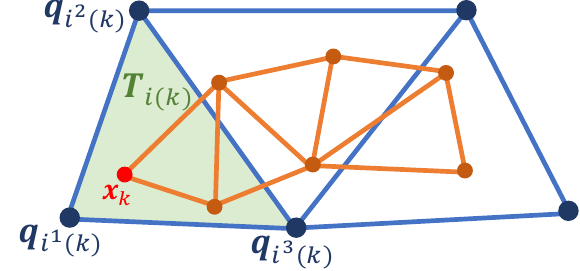}
  \caption{A simple 2-dimensional case illustration. The blue mesh is the embedding for the orange collision mesh. We denote the red vertex of the collision mesh by $\vf{x}_k$, the associated embedding triangle of $\vf{x}_k$ is $T_{i(k)}$ which is highlighted by the green area. In a 3-dimensional space it should be an embedding tetrahedron as we stated in Sec. \ref{sec:embedding_impl}. The vertex positions of $T_{i(k)}$ are $\vf{q}_{i^j(k)}, j \in \mathbb{N}, 1 \leq j \leq d + 1$, where $d$ is the world-space dimension. }
  \label{fig:subspace}
  \vspace{-5mm}
\end{figure}

We assume each object admits a high-resolution triangle surface collision mesh $M_{\text{col}} = (X, S)$ where $X \subset \{\vf{x}_1, \vf{x}_2, ..., \vf{x}_{N_v}\}$ and $S$ is a collection of triplets recording the indices of the three vertices of each triangle. As shown in Fig. \ref{fig:subspace}, a low-resolution embedding tetrahedral mesh $M_{\text{emb}} = (\{\vf{q}_1, \vf{q}_2, ..., \vf{q}_{N_s}\}, T)$ is associated with $M_{\text{col}}$ such that $\forall \vf{x}_k \in X,$ there exists a unique tetrahedral $T_{i(k)} = [i^1(k), i^2(k), i^3(k), i^4(k)] \in T$ with vertex positions $\vf{q}_{i^1(k)^1}, \vf{q}_{i^2(k)}, \vf{q}_{i^3(k)},$ and $\vf{q}_{i^4(k)}$ containing $\vf{x}_k$ in its internal volume. At time $t=0,$ for each vertex $\vf{x}_k$, we compute the barycentric weights $\{\omega^k_{j}\}_{j=1}^4$ as 
\begin{equation}
    \vf{x}_k^0 = \sum_{j=1}^4 \omega^k_{j} \vf{q}^0_{i^j(k)}.
\end{equation}
We can then express $\vf{x}_k$ at arbitrary time step $t_n$ as 
\begin{equation}
    \vf{x}_k^n = \sum_{j=1}^4 \omega^k_{j} \vf{q}^n_{i^j(k)}.
    \label{eqn:embedding_mapping}
\end{equation}
The reduced generalized coordinates vector $\mf{q}$ is thus defined by stacking all vertex positions in $M_{\text{emb}}$ as $\mf{q} = [\vf{q}_1^T, \vf{q}_2^T, ..., \vf{q}_{N_s}^T]^T.$ Eq. (\ref{eqn:embedding_mapping}) provides the definition of our chosen embedding mapping $\phi(\mf{q})$.

With this choice of $\phi(\mf{q})$, the elastic energy %
$\Phi^{\mf{q}}(\mf{q})=\Phi^{\mf{x}}(\phi(\mf{q}))$ can be directly computed on ${M}_{\text{emb}}$ as 
\begin{align}
\Phi^{\mf{x}}(\phi(\mf{q})) &= \int_{\Omega} 
\Psi(\mf{F}(\vf{x})) d\vf{x} \approx \sum_{k} \Psi(\mf{F}_{k}^{\mf{x}}) V_k \nonumber \\
&= \sum_{k} \Psi(\mf{F}_{T_{i(k)}}) V_k = \sum_{k} \sum_j \Psi(\mf{F}_{T_j})V_k I_{\{k: i(k) = j\}}(k) \nonumber \\ 
&= \sum_{j}\Psi(\mf{F}_{T_j}) \sum_{k} V_k I_{\{k: i(k) = j\}}(k) = \sum_{j} \Psi(\mf{F}_{T_j})V_{T_j},
\end{align}
where $\Omega$ represents the internal volume of $M_{\text{col}}$ at $t=0$,  $\mf{F}(\vf{x}^0)=\frac{\partial \vf{x}}{\partial \vf{x}^0} \in \mathbb{R}^{3 \times 3}$ is the deformation gradient at $\vf{x}$,  $\mf{F}_{k}^{\mf{x}}=\mf{F}(\vf{x}_k^0)$ is its value at $\vf{x}_k$. Note that $\mf{F}_{k}^{\mf{x}}=\mf{F}_{T_{i(k)}}$, where $\mf{F}_{T_{i(k)}}=\mf{D}_{i(k)}(\mf{D}^0_{i(k)})^{-1}\in \mathbb{R}^{3\times 3}$ is the deformation gradient of the embedding tetrahedra $T_{i(k)}$, $ \mf{D}_{i(k)}^0 = \mf{D}_{i(k)}\Bigr\rvert_{t=0}$, and 
\begin{equation*}
    \mf{D}_{i(k)}=\left[\vf{q}_{i^2(k)} - \vf{q}_{i^1(k)}, \vf{q}_{i^3(k)} - \vf{q}_{i^1(k)}, \vf{q}_{i^4(k)} - \vf{q}_{i^1(k)}\right]\in \mathbb{R}^{3\times 3}.
\end{equation*}
$\Psi(\mf{F})$ is an elastic energy density function of deformation gradient $\mf{F}$; $V_k$ is per-vertex volume of $\vf{x}_k$, and $V_{T_i} = \sum_{k} V_k I_{\{k: i(k) = i\}}(k).$
The elasticity energy can then be computed on ${M}_{\text{emb}}$ as if ${M}_{\text{emb}}$ has the same constitutive model as ${M}_{\text{col}}$ and each tetrahedron $T_i$ of ${M}_{\text{emb}}$ has a volume of $V_{T_i}$. 

In addition, the potential energy $V(\mf{q})$ also includes an external force term $E_{\text{ext}}(\mf{q})$. Denote the external force in full space by 
$\mf{f}_{\text{ext}} \in \mathbb{R}^{3N_v}.$ It follows that
\begin{equation}
    E_{\text{ext}}(\mf{q})=\mf{f}_{\text{ext}}^T \mf{x} = \mf{f}_{\text{ext}}^T(\mf{Jq}) = (\mf{J}^T\mf{f}_{\text{ext}})^T \mf{q}=(\mf{f}^{\mf{q}}_{\text{ext}})^T\mf{q},
\end{equation}
where $\mf{f}^{\mf{q}}_{\text{ext}}=\mf{J}^T\mf{f}_{\text{ext}}$ is the generalized external force.  

Notice that the gradient and hessian in the subspace $Q$ needed in Newton iterations can be conveniently obtained as 
\begin{equation}
    \nabla_{\mf{q}}E = \mf{J}^T\nabla_{\mf{x}}E, \text{ and } \nabla^2_{\mf{q}}E = \mf{J}^T \nabla^2_{\mf{x}}E \mf{J}.
\end{equation}

So far, we have derived a concrete simulation pipeline under a subspace simulation framework and a variational framework. 
The full-space \ac{ipc} is also contained in our framework as a special case where $Q=\mathbb{R}^{3N_{v}}$ and $\phi$ is the identity mapping. Another interesting case is if we choose a tetrahedron containing a collision mesh ${M}_{\text{col}}$ as the simulation mesh, where the collision mesh can only have affine deformation in our simulation algorithm. Its elastic energy will be $\Phi(\mf{q}) = V_{{M}_{\text{col}}}\Psi(\mf{F})$, where $V_{{M}_{\text{col}}}$ is the volume of ${M}_{\text{col}}$, $\mf{F}$ is the deformation gradient corresponding to the affine deformation of ${M}_{\text{col}}$. By taking $\Psi(\mf{F})=\kappa \Vert \mf{F}\mf{F}^T - \mf{I}\Vert_F^2$ where $\kappa > 0$ is a stiffness parameter, we can see this special case is exactly equivalent to the \ac{abd}~\cite{lan2022affine} simulation. In this sense, our simulation framework unifies full-space \ac{ipc} and \ac{abd} simulation algorithms. By choosing a simulation mesh finer than a single tetrahedron but coarser than the high-resolution collision mesh ${M}_{\text{col}}$, we can benefit from both high accuracy and efficiency from the trade-off.

\section{EXPERIMENTS}

We quantitatively evaluate the convergence, frictional contact resolution accuracy, and performance of our proposed method in a grasping experiment using bubble grippers. Further, we simulate a more challenging scenario of placing a thin plate to demonstrate the robustness and intersection-free guarantee of our approach. All simulations are performed on Intel(R) Core(TM) i9-14900KF (16-core). Gravitational acceleration is set to $g=9.81 \text{ m}\cdot \text{s}^{-2}$ in both experiments.

\subsection{Grasping a teddy bear by a soft bubble gripper}
\label{sec:bowl_rack}
\begin{figure}
\includegraphics[width=1.0\linewidth]{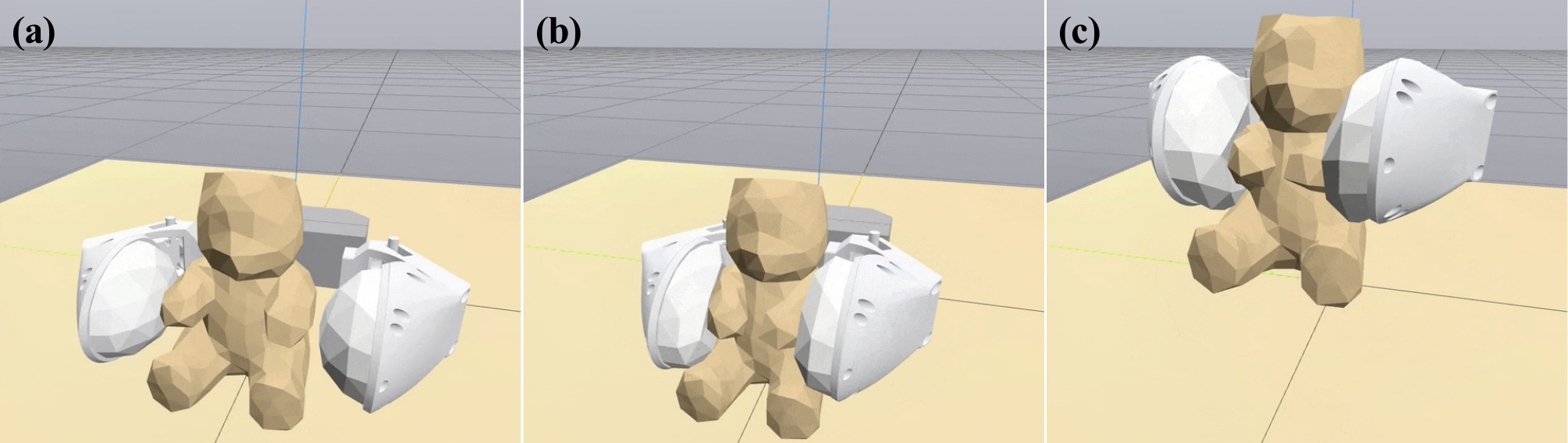}
  \caption{We simulate grasping a soft teddy bear with a soft bubble gripper. The process contains 3 steps: \textit{(i)}. \textbf{Grasping}: Moving the bubbles toward the teddy bear to grasp it, moving from \textbf{(a)} to \textbf{(b)}). \textit{(ii)}. \textbf{Lifting}: Lifting the teddy bear vertically off the ground, moving from \textbf{(b)} to \textbf{(c)}. \textit{(iii)}. \textbf{Holding}: The bubble gripper remains stationary, holding the teddy bear as in \textbf{(c)}. 
  }
  \vspace{-3mm}\label{fig:bubble_gripper_scene}
\end{figure}

In this experiment, a pair of soft bubbles~\cite{kuppuswamy2020soft} attached to gripper fingers is used to grasp a deformable teddy bear, as shown in Fig. \ref{fig:bubble_gripper_scene}. In addition to the full-space IPC method~\cite{li2020incremental}, we choose the method proposed in~\cite{drake_fric_0, drake_fric_1} and implemented in  Drake \cite{drake} as baselines. Drake is chosen for its accurate frictional contact handling, support for soft gripper modeling, and open-source accessibility.

In our method, the bubbles are modeled as regular soft bodies as in full-space IPC since each bubble contains only 67 vertices, while the teddy bear is modeled with the proposed Embedded IPC. For these methods, we set the contact parameters to be $\kappa=10^4 \text{kg}\cdot \text{s}^{-2}$, $\hat{d}=10^{-3}$m and $\epsilon_{v}=10^{-3}$m/s. Both the bubbles and the teddy bear are modeled with linear corotational elasticity model. The bubbles have Young's modulus $E_{\text{bubble}}=10^4 \text{ Pa}$, Poisson's ratio $\nu_{\text{bubble}}=0.45$ and mass density of $\rho_{\text{bubble}}=10 \text{ kg} \cdot \text{m}^{-3}$; the teddy bear has Young's modulus $E_{\text{teddy}}=5\times10^{4} \text{ Pa}$, Poisson's ratio $\nu_{\text{teddy}}=0.45$ and mass density of $\rho_{\text{teddy}}=10^{3} \text{ kg}\cdot \text{m}^{-3}$. The friction coefficient is set as $\mu_{\text{friction}}=1.0$.

In the simulation, the bubble gripper first grasps the teddy bear on the ground by compressing it, then lifts it upwards, followed by a final stop; see Fig. \ref{fig:bubble_gripper_scene}. The squeezing and lifting phase each lasts 1.5 seconds, followed by a 1-second pause, resulting in a total simulation time of 4 seconds. We use position control in simulations, where a specific set of vertices on the bubble surfaces are constrained to follow a prescribed motion. Since the performance of our method depends on the resolution of the constructed subspace, we present results for both medium-resolution and low-resolution meshes used in the subspace construction. The original high-resolution teddy bear tetrahedral mesh contains 410 vertices and 1207 cells, while the medium-resolution one contains 173 vertices and 533 cells, and the low-resolution one has 34 vertices and 61 cells. The complete motion is shown in the supplemental video.

\subsubsection{Contact Force Analysis}
\begin{figure}
\centering
\includegraphics[width=1.0\linewidth]{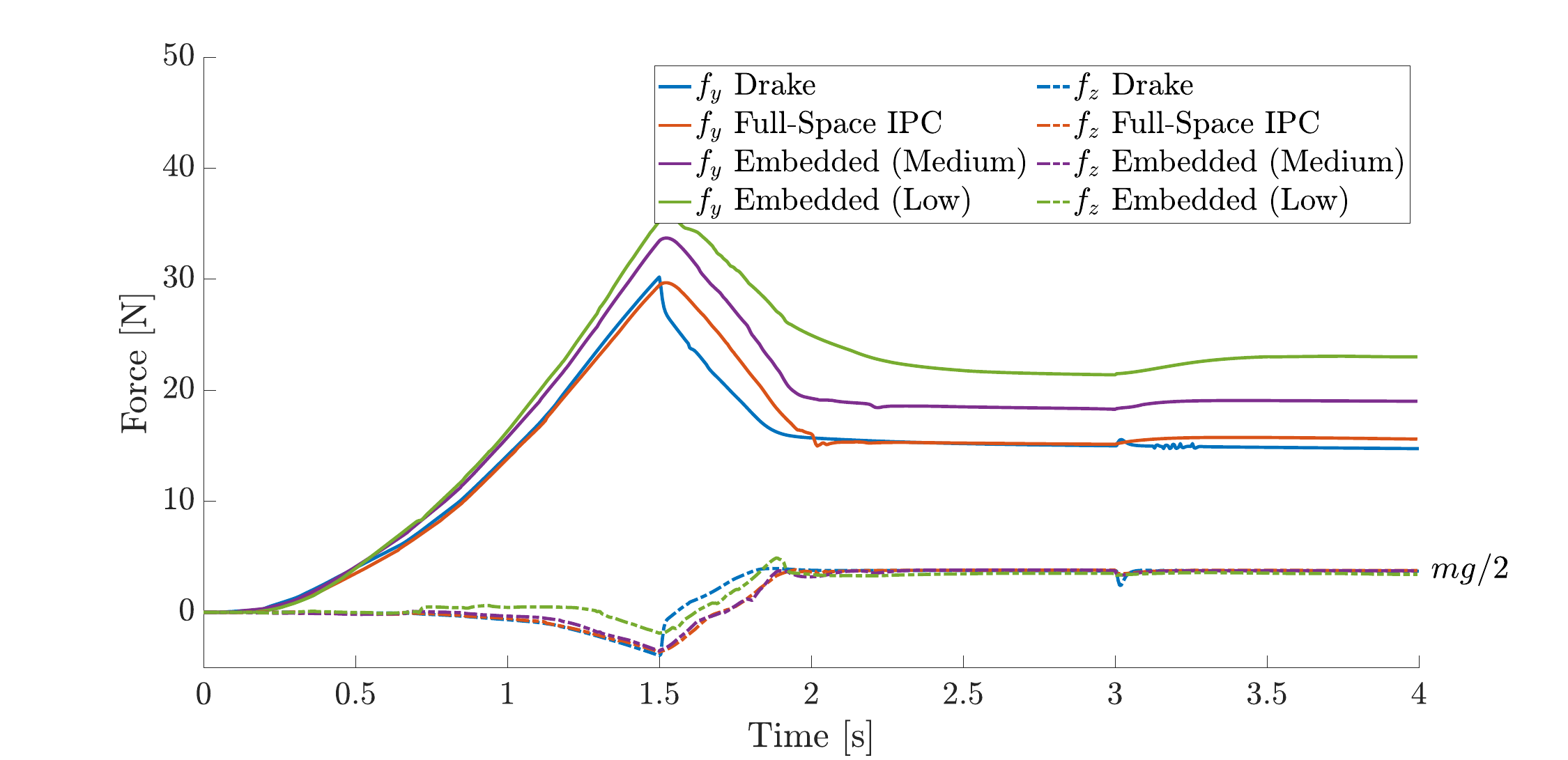}
\caption{Contact forces between the left deformable bubble and the teddy bear. The simulation has a time step size of $h = 0.005\text{s}$.}
\label{fig:bubble_gripper_forces}
\end{figure}

\begin{figure}
\centering
\vspace{-3mm}
\includegraphics[width=1.0\linewidth]{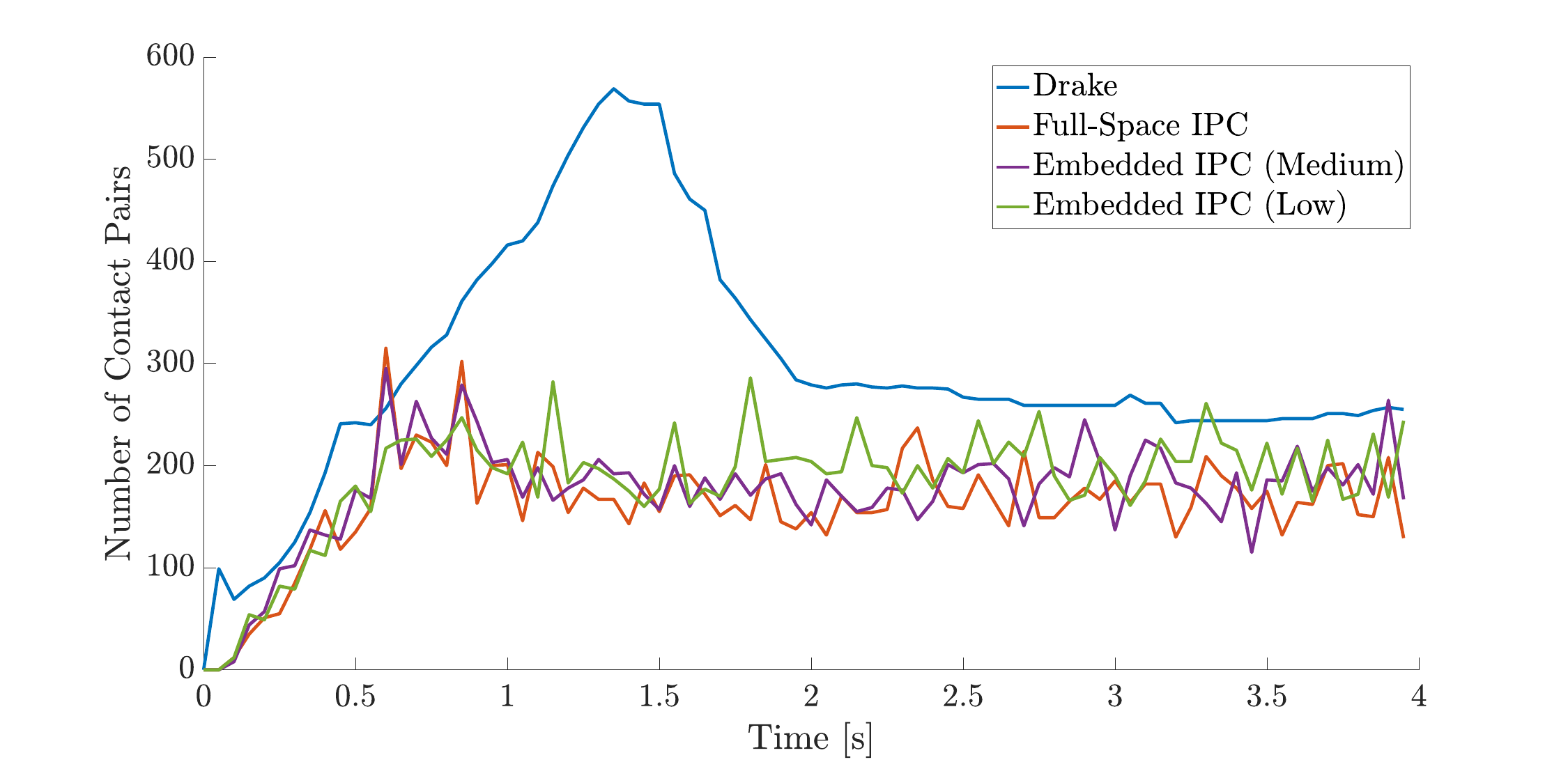}
\caption{Number of contacts as a function of time.  The simulation has a time-step of $h = 0.005s$ as in Fig. \ref{fig:bubble_gripper_forces}.}
\label{fig:bubble_gripper_contact_number}
\vspace{-5mm}
\end{figure}

Fig. \ref{fig:bubble_gripper_forces} plots the contact forces for each algorithm. A sudden change in the derivative of the force magnitude is observed at $t=1.5\text{s}$ as the grippers start to lift the teddy bear. At $t = 3.0\text{s},$ the grippers are set to still, causing the teddy bear to bounce momentarily due to elasticity and inertia. The force curves of the full-space IPC align closely with those of Drake, with acceptable discrepancies due to distinct contact models. When Rayleigh damping is applied, IPC introduces more dissipation compared to Drake, which needs further investigation and is out of the scope of the current work. Frictional forces in the z-direction converge to the analytical solution for all four simulations. In the squeezing direction, a finer simulation mesh leads to a smaller contact force $f_y$. This is because a finer discretization can capture more detailed deformation, whereas a coarser mesh may lead to more global deformation and thus larger contact forces in the y-direction. The number of contact pairs during the simulations are plotted in Fig. \ref{fig:bubble_gripper_contact_number}. The simulation with Drake has more contact pairs, likely due to its reliance on slight penetrations to resolve contacts.

\begin{figure} [b]
\centering
\vspace{-6mm}
\includegraphics[width=1.0\linewidth]{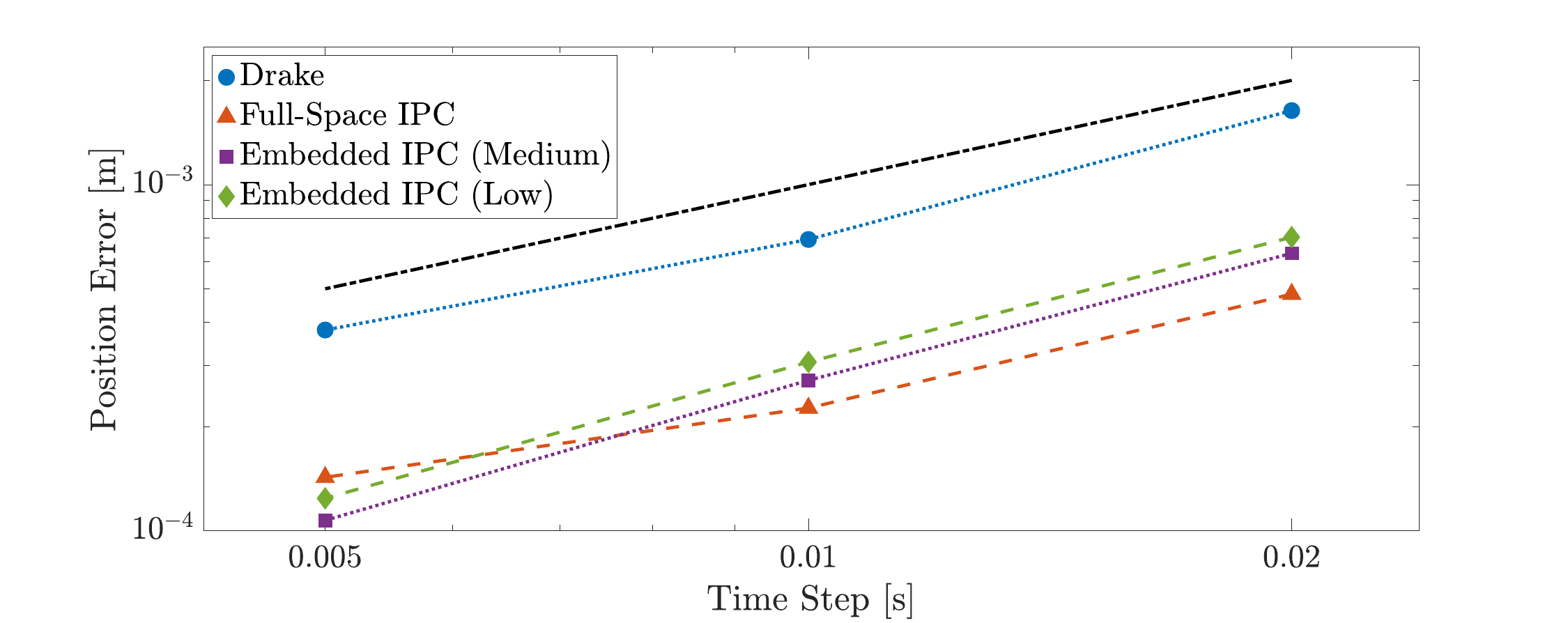}
\caption{Convergence with time step of each method. The dashed black line is a first order reference. All the methods show a linear rate of convergence with regard to the time step size $h$. }
\label{fig:bubble_gripper_error_timestep}
\vspace{-2.0mm}
\end{figure}

\begin{figure}
\centering
\includegraphics[width=1.0\linewidth]{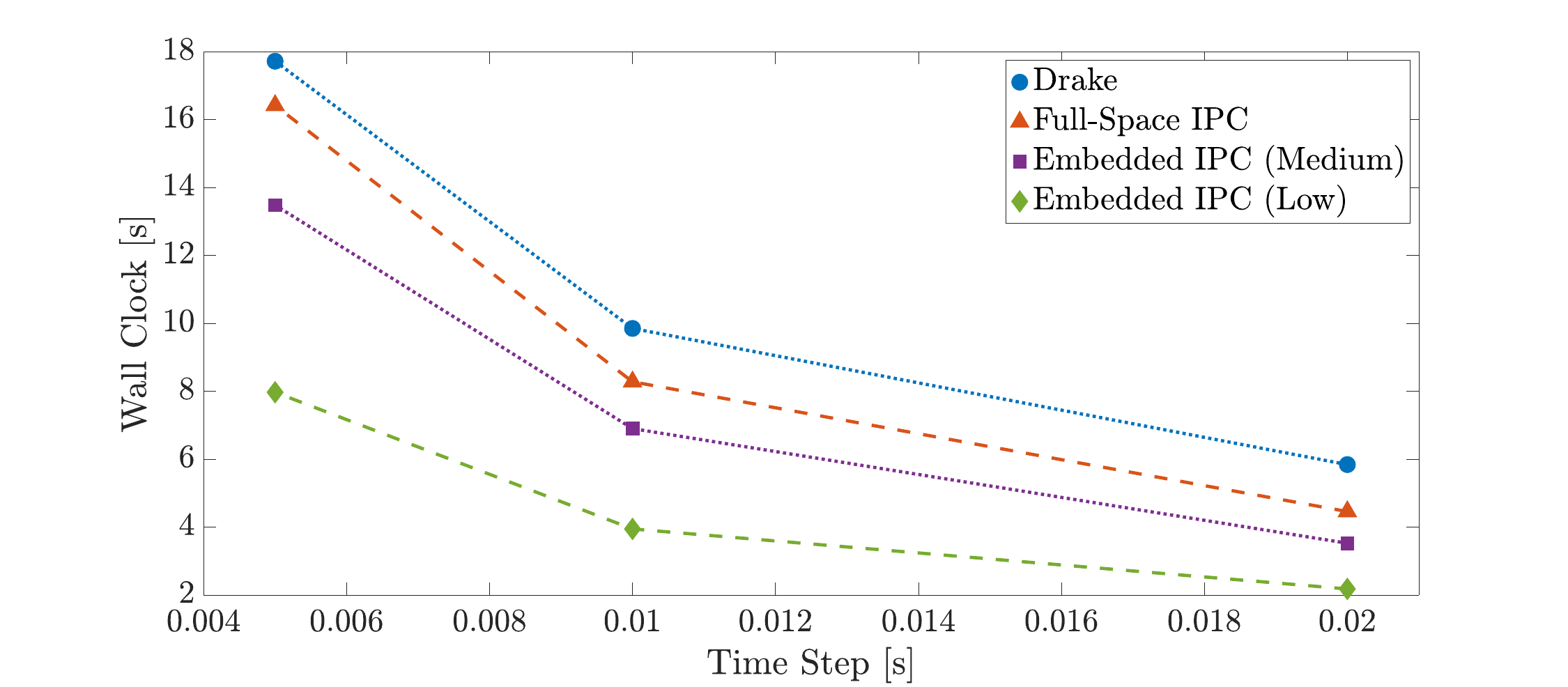}
\caption{Wall-clock time as a function of time step size of each method. Our method is faster than all other baseline methods. }
\label{fig:bubble_gripper_timestep_wall-clock}
\vspace{-2.0mm}
\end{figure}

\begin{figure}
\centering
\includegraphics[width=1.0\linewidth]{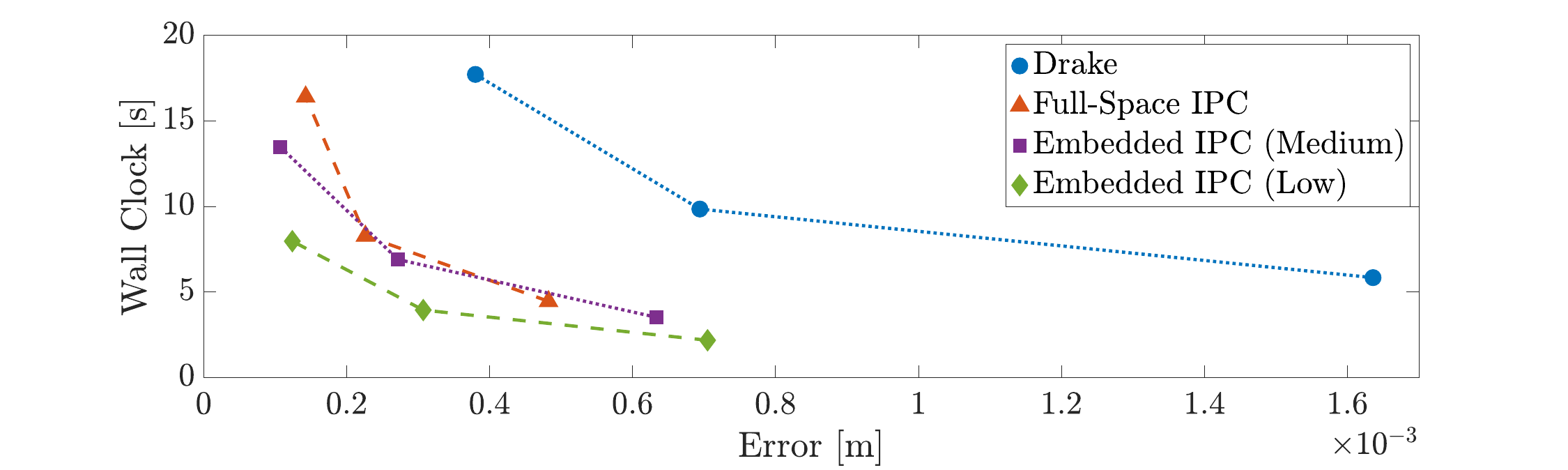}
\caption{This plot is made from the accuracy (errors) measured for Fig. \ref{fig:bubble_gripper_error_timestep} along with the costs (wall-clock) reported in Fig. \ref{fig:bubble_gripper_timestep_wall-clock}, for the corresponding time step sizes. Our method converges significantly faster to its own reference solution than baseline methods. }
\label{fig:bubble_gripper_error_wall-clock}
\vspace{-2.0mm}
\end{figure}

\subsubsection{Convergence and Efficiency Study}
We compare the convergence and efficiency of our method with benchmark methods by running simulations with various time step sizes $h: 0.005\text{s}, 0.01\text{s} \text{ and } 0.02\text{s}$. The ground truth solution is approximated by running each algorithm with a small time step of $h_\text{ref}=5\times10^{-4}\text{s}$. Since each method adopts different approximations, a reference solution $\mf{x}_{\text{ref}}$ is computed for each to ensure a fair comparison. We choose a time-averaged error metric $\mathcal{E}(h)$ defined as $\mathcal{E}(h) = \sqrt{\frac{1}{\floor*{T/h}}\sum_{i=1}^{\floor*{T/h}} \frac{1}{N_v} \Vert \mf{x}_{h}(ih) - \mf{x}_{\text{ref}}(ih)\Vert^2}$, where $\mf{x}_{h}(t)$ denotes vertex positions in the solution obtained with time step size $h.$ $T$ is the total simulation time. We plot the wall-clock running times and position errors against time step sizes in Fig. \ref{fig:bubble_gripper_error_timestep} and Fig. \ref{fig:bubble_gripper_timestep_wall-clock}.

All methods have an $O(h)$ convergence rate, as shown in Fig. \ref{fig:bubble_gripper_error_timestep}. 
Fig. \ref{fig:bubble_gripper_timestep_wall-clock} reveals that our method runs faster at the same time step size. We also plot the position error and wall-clock runtime in three runs for each method using different time step sizes 
in Fig. \ref{fig:bubble_gripper_error_wall-clock} to as a way to compare performance for a given accuracy. This acceleration partially comes from the multi-threaded parallel implementation of our method and the full-space \ac{ipc} method, whereas Drake runs on a single thread. In our method, enabling multi-threaded will speed up the overall performance by $1.4\times$ compared to single-threaded. This modest acceleration is primarily due to the small scale of the problem. Compared with vanilla full-space \ac{ipc}, our method runs $2.0\times$ faster using the low-resolution embedding mesh, showing its high efficiency with little compromise on accuracy. 
In this experiment, with a time step size of $h=0.02\text{s}$ and the low-resolution subspace embedding, our method runs at $1.8\times$ real-time rate, enabling interactive capability with non-penetration guarantee.

\subsection{Placing a plate on a dish rack}
\begin{figure}
\centering
\includegraphics[width=0.9\linewidth]{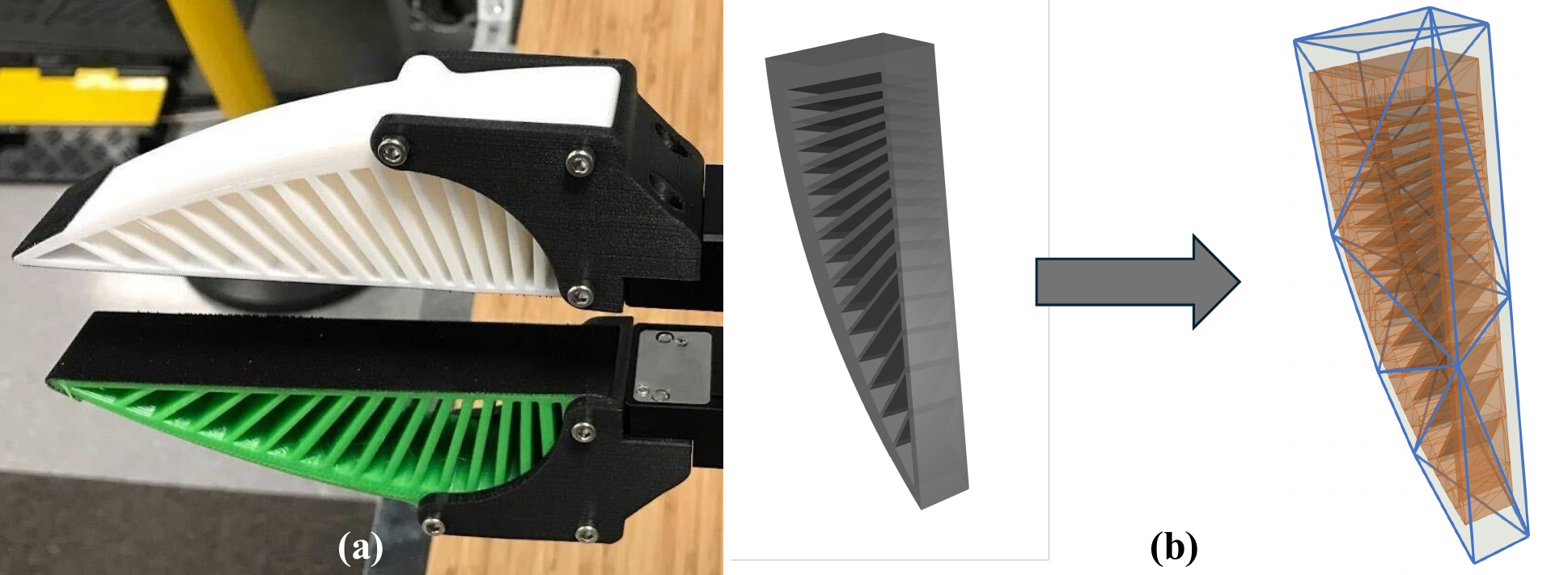}
\caption{The modeling of the deformable FinRay gripper used in Sec. \ref{sec:bowl_rack}. \textbf{(a)} The real gripper used for trajectories recording. \textbf{(b)} Our embedding modeling. The original mesh and embedding mesh are shown in orange and blue respectively, as in Fig. \ref{fig:subspace}.}
\label{fig:finray_gripper}
\vspace{-5mm}
\end{figure}

We present a more challenging task where a deformable FinRay gripper (see Fig. \ref{fig:finray_gripper}) \cite{chi2024universal} is used to grasp a thin plate and place it inside the dish rack. Two trajectories from real world experiments are recorded. We simulate this scene with our and baseline methods (\cite{castro2022unconstrained} in Drake and PhysX~\cite{nvidia_physx} in Isaac Sim) according to the trajectories.

\begin{figure} [b]
\centering
\vspace{-2mm}
\includegraphics[width=1.0\linewidth]{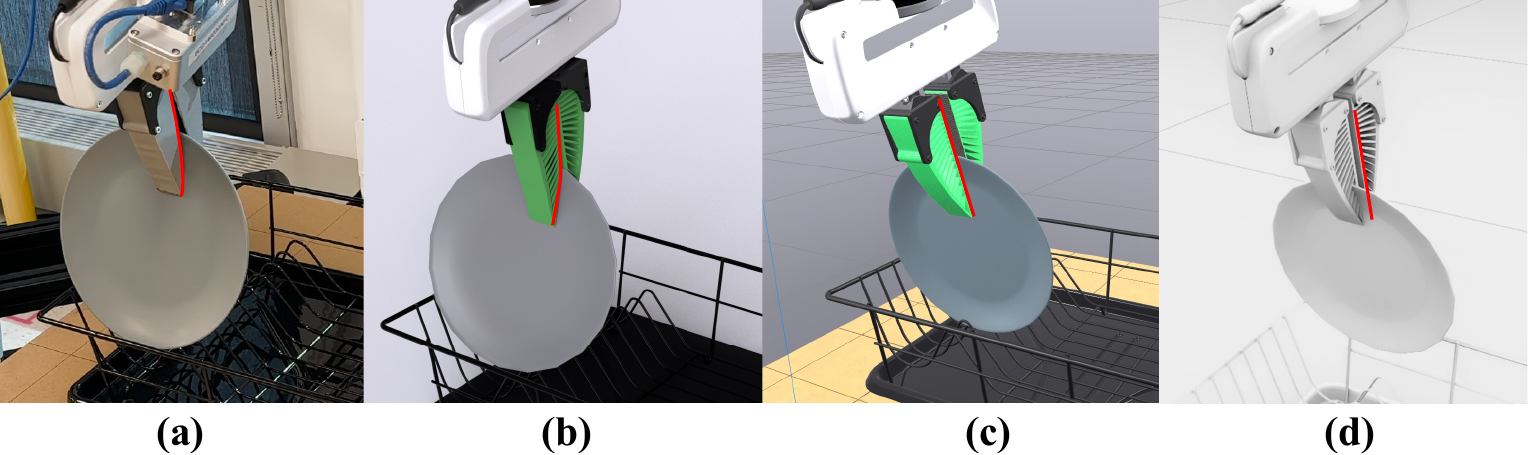}
\vspace{-4mm}
\caption{A FinRay gripper grasping a plate in the real world \textbf{(a)}, in our simulator \textbf{(b)}, Drake \textbf{(c)} and Isaac Sim \textbf{(d)}. Our method effectively captures the soft gripper deformation. Meanwhile, in Drake, the gripper links have no deformation as they are modeled as rigid bodies due to efficiency issues. Isaac Sim shows undesired penetration artifacts and fails to replay the two trajectories, resulting in unstable simulations. More Isaac Sim results can be found in our supplementary material.}
\label{fig:dish_rack_finray_deformation}
\end{figure}

The dish rack and the plate are modeled as rigid bodies in Drake and Isaac Sim. In our method and the full-space IPC, they are modeled as stiff affine bodies with Young's modulus of $E_\text{rigid}=10^7 \text{Pa}.$ The FinRay gripper has anisotropic physical properties due to its cut-out pattern. Therefore, it bends easily when the black strip is pressed vertically (see Fig. \ref{fig:finray_gripper} \textbf{(a)}), albeit its high stiffness. This particular deformation mode enables the gripper to conform to the shape of the manipuland, thus increasing contact area and improving grasp stability. In Drake, for real-time performance, the gripper in this setup is modeled using a compliant model of contact surfaces \cite{hydroelastic, drake_hydroelastic}. While this contact model has proved useful in practice, it is an approximation that does not resolve the deformations of the real gripper. Isaac Sim shows undesired penetration artifacts and fails to replay the two trajectories due to the unstable simulation since the constraint-based method cannot converge in the tight time budget and leads to deep penetrations. By contrast, using a linear corotated elasticity model, our method and the full-space IPC directly model these links as deformable objects. As shown in Fig. \ref{fig:dish_rack_finray_deformation}, compared to Drake, our method generates bending deformations in the soft gripper that are highly consistent with real-world results, outperforming Drake in terms of realism with comparable computational efficiency. 

\begin{figure}
\centering
\includegraphics[width=0.7\linewidth]{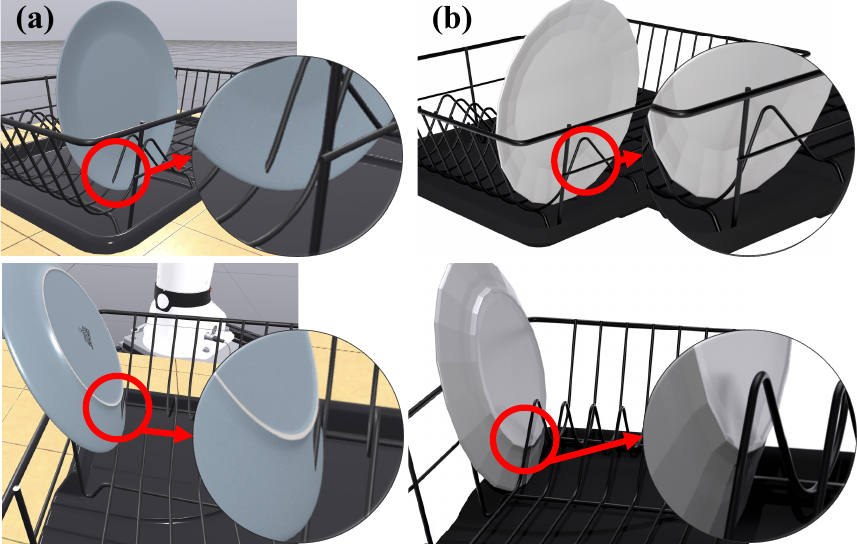}
\caption{Snapshots during contact-rich simulations in our method \textbf{(a)} and Drake \textbf{(b)} where we use a FinRay gripper to grasp and place a rigid plate into a dish rack slot. In drake, penetrations occur due to intensive contacts between the plate and the thin wires of the dish rack wireframe. While our method show no penetrations as it provides an intersection-free gauarantee. }
\vspace{-4mm}
\label{fig:dish_rack_penetration_results}
\end{figure}

Derived from full-space IPC methods, our method also inherits the strict intersection-free guarantee property. The tiny thickness of the wires and the plate, as well as the complex geometry of the rack, all poses challenge to contact solving. As demonstrated in Fig. \ref{fig:dish_rack_penetration_results}, severe penetrations can be observed in Drake's results when complex contacts occur, while our method provides intersection-free results consistent with real-world results, despite intensive contacts.

We note that inconsistent plate states between simulations and  real world results are noticeable both in our simulation and baseline simulations, mainly due to calibration errors, measurement inaccuracies, imprecise physical parameters, contact model approximations, and misaligned motion control parameters used in simulations. Due to the challenge of contact physics, these errors accumulate and amplify over time, resulting in simulation results inconsistent with real-world ones. 

\section{CONCLUSIONS}
We propose \emph{Embedded IPC}, an intersection-free method that leverages model reduction to enable simulation of deformable objects at real time rates. We performed a time step convergence study demonstrating our method is first order in time step size. Moreover, we show \emph{Embedded IPC} runs at real time rates while still providing accurate solutions when compared with other simulation baselines. Embedded \ac{ipc} generates intersection-free solutions, enabling users to simulate complex contact-rich manipulation tasks that are challenging to replicate in other real-time simulators. While intersection free, IPC introduces \emph{action at a distance} within a thin compliant layer around objects. Further research is needed to assess the practical implications of these approximations for robotic applications. Moreover, our method is prone to locking issues when using a small subspace dimension to simulate large deformations. To address this, we intend to encode various deformation modes into the subspace in the future. In this paper, we only deal with volumetric soft bodies. We plan to choose a proper subspace for co-dimensional objects and integrate them with the current simulation algorithms to support a broader range of robotic applications and scenarios. Currently, we use heuristic ways to construct embedding mesh for each object. Developing an embedding mesh generation algorithm will enhance the automation of this process. We leave these for future work.  

\bibliographystyle{IEEEtran}
\bibliography{com}

\begin{thebibliography}{10}
\providecommand{\url}[1]{#1}
\csname url@samestyle\endcsname
\providecommand{\newblock}{\relax}
\providecommand{\bibinfo}[2]{#2}
\providecommand{\BIBentrySTDinterwordspacing}{\spaceskip=0pt\relax}
\providecommand{\BIBentryALTinterwordstretchfactor}{4}
\providecommand{\BIBentryALTinterwordspacing}{\spaceskip=\fontdimen2\font plus
\BIBentryALTinterwordstretchfactor\fontdimen3\font minus \fontdimen4\font\relax}
\providecommand{\BIBforeignlanguage}[2]{{%
\expandafter\ifx\csname l@#1\endcsname\relax
\typeout{** WARNING: IEEEtran.bst: No hyphenation pattern has been}%
\typeout{** loaded for the language `#1'. Using the pattern for}%
\typeout{** the default language instead.}%
\else
\language=\csname l@#1\endcsname
\fi
#2}}
\providecommand{\BIBdecl}{\relax}
\BIBdecl

\bibitem{schmitt2018soft}
F.~Schmitt, O.~Piccin, L.~Barb{\'e}, and B.~Bayle, ``Soft robots manufacturing: A review,'' \emph{Frontiers in Robotics and AI}, vol.~5, p.~84, 2018.

\bibitem{zhu2022challenges}
J.~Zhu, A.~Cherubini, C.~Dune, D.~Navarro-Alarcon, F.~Alambeigi, D.~Berenson, F.~Ficuciello, K.~Harada, J.~Kober, X.~Li \emph{et~al.}, ``Challenges and outlook in robotic manipulation of deformable objects,'' \emph{IEEE Robotics \& Automation Magazine}, vol.~29, no.~3, pp. 67--77, 2022.

\bibitem{agarwal2021simulation}
A.~Agarwal, T.~Man, and W.~Yuan, ``Simulation of vision-based tactile sensors using physics based rendering,'' in \emph{2021 IEEE International Conference on Robotics and Automation (ICRA)}.\hskip 1em plus 0.5em minus 0.4em\relax IEEE, 2021, pp. 1--7.

\bibitem{ding2021sim}
Z.~Ding, Y.-Y. Tsai, W.~W. Lee, and B.~Huang, ``Sim-to-real transfer for robotic manipulation with tactile sensory,'' in \emph{2021 IEEE/RSJ International Conference on Intelligent Robots and Systems (IROS)}.\hskip 1em plus 0.5em minus 0.4em\relax IEEE, 2021, pp. 6778--6785.

\bibitem{kadian2020sim2real}
A.~Kadian, J.~Truong, A.~Gokaslan, A.~Clegg, E.~Wijmans, S.~Lee, M.~Savva, S.~Chernova, and D.~Batra, ``Sim2real predictivity: Does evaluation in simulation predict real-world performance?'' \emph{IEEE Robotics and Automation Letters}, vol.~5, no.~4, pp. 6670--6677, 2020.

\bibitem{todorov2012mujoco}
E.~Todorov, T.~Erez, and Y.~Tassa, ``Mujoco: A physics engine for model-based control,'' in \emph{2012 IEEE/RSJ International Conference on Intelligent Robots and Systems}.\hskip 1em plus 0.5em minus 0.4em\relax IEEE, 2012, pp. 5026--5033.

\bibitem{makoviychuk2021isaac}
V.~Makoviychuk, L.~Wawrzyniak, Y.~Guo, M.~Lu, K.~Storey, M.~Macklin, D.~Hoeller, N.~Rudin, A.~Allshire, A.~Handa \emph{et~al.}, ``Isaac gym: High performance gpu-based physics simulation for robot learning,'' \emph{arXiv preprint arXiv:2108.10470}, 2021.

\bibitem{castro2022unconstrained}
A.~M. Castro, F.~N. Permenter, and X.~Han, ``An unconstrained convex formulation of compliant contact,'' \emph{IEEE Transactions on Robotics}, vol.~39, no.~2, pp. 1301--1320, 2022.

\bibitem{han2023convex}
X.~Han, J.~Masterjohn, and A.~Castro, ``A convex formulation of frictional contact between rigid and deformable bodies,'' \emph{IEEE Robotics and Automation Letters}, 2023.

\bibitem{faure2012sofa}
F.~Faure, C.~Duriez, H.~Delingette, J.~Allard, B.~Gilles, S.~Marchesseau, H.~Talbot, H.~Courtecuisse, G.~Bousquet, I.~Peterlik \emph{et~al.}, ``Sofa: A multi-model framework for interactive physical simulation,'' \emph{Soft tissue biomechanical modeling for computer assisted surgery}, pp. 283--321, 2012.

\bibitem{duriez2013control}
C.~Duriez, ``Control of elastic soft robots based on real-time finite element method,'' in \emph{2013 IEEE international conference on robotics and automation}.\hskip 1em plus 0.5em minus 0.4em\relax IEEE, 2013, pp. 3982--3987.

\bibitem{wang2023dexgraspnet}
R.~Wang, J.~Zhang, J.~Chen, Y.~Xu, P.~Li, T.~Liu, and H.~Wang, ``Dexgraspnet: A large-scale robotic dexterous grasp dataset for general objects based on simulation,'' in \emph{2023 IEEE International Conference on Robotics and Automation (ICRA)}.\hskip 1em plus 0.5em minus 0.4em\relax IEEE, 2023, pp. 11\,359--11\,366.

\bibitem{zhao2020sim}
W.~Zhao, J.~P. Queralta, and T.~Westerlund, ``Sim-to-real transfer in deep reinforcement learning for robotics: a survey,'' in \emph{2020 IEEE symposium series on computational intelligence (SSCI)}.\hskip 1em plus 0.5em minus 0.4em\relax IEEE, 2020, pp. 737--744.

\bibitem{li2020incremental}
M.~Li, Z.~Ferguson, T.~Schneider, T.~R. Langlois, D.~Zorin, D.~Panozzo, C.~Jiang, and D.~M. Kaufman, ``Incremental potential contact: intersection-and inversion-free, large-deformation dynamics.'' \emph{ACM Trans. Graph.}, vol.~39, no.~4, p.~49, 2020.

\bibitem{kim2022ipc}
C.~M. Kim, M.~Danielczuk, I.~Huang, and K.~Goldberg, ``Ipc-graspsim: Reducing the sim2real gap for parallel-jaw grasping with the incremental potential contact model,'' in \emph{2022 International Conference on Robotics and Automation (ICRA)}.\hskip 1em plus 0.5em minus 0.4em\relax IEEE, 2022, pp. 6180--6187.

\bibitem{du2024intersection}
W.~Du, S.~Yao, X.~Wang, Y.~Xu, W.~Xu, and C.~Lu, ``Intersection-free robot manipulation with soft-rigid coupled incremental potential contact,'' \emph{IEEE Robotics and Automation Letters}, 2024.

\bibitem{du2024tacipc}
W.~Du, W.~Xu, J.~Ren, Z.~Yu, and C.~Lu, ``Tacipc: Intersection-and inversion-free fem-based elastomer simulation for optical tactile sensors,'' \emph{IEEE Robotics and Automation Letters}, 2024.

\bibitem{chen2024general}
W.~Chen, J.~Xu, F.~Xiang, X.~Yuan, H.~Su, and R.~Chen, ``General-purpose sim2real protocol for learning contact-rich manipulation with marker-based visuotactile sensors,'' \emph{IEEE Transactions on Robotics}, 2024.

\bibitem{fernandez2024stark}
J.~A. Fernández-Fernández, R.~Lange, S.~Laible, K.~O. Arras, and J.~Bender, ``Stark: A unified framework for strongly coupled simulation of rigid and deformable bodies with frictional contact,'' in \emph{2024 IEEE International Conference on Robotics and Automation (ICRA)}, 2024, pp. 16\,888--16\,894.

\bibitem{chi2024universal}
C.~Chi, Z.~Xu, C.~Pan, E.~Cousineau, B.~Burchfiel, S.~Feng, R.~Tedrake, and S.~Song, ``Universal manipulation interface: In-the-wild robot teaching without in-the-wild robots,'' \emph{arXiv preprint arXiv:2402.10329}, 2024.

\bibitem{bathe2007finite}
K.~J. Bathe, ``Finite element method,'' \emph{Wiley encyclopedia of computer science and engineering}, pp. 1--12, 2007.

\bibitem{otaduy2009implicit}
M.~A. Otaduy, R.~Tamstorf, D.~Steinemann, and M.~Gross, ``Implicit contact handling for deformable objects,'' in \emph{Computer Graphics Forum}, vol.~28, no.~2.\hskip 1em plus 0.5em minus 0.4em\relax Wiley Online Library, 2009, pp. 559--568.

\bibitem{baraff1994fast}
D.~Baraff, ``Fast contact force computation for nonpenetrating rigid bodies,'' in \emph{Proceedings of the 21st annual conference on Computer graphics and interactive techniques}, 1994, pp. 23--34.

\bibitem{stewart2000rigid}
D.~E. Stewart, ``Rigid-body dynamics with friction and impact,'' \emph{SIAM review}, vol.~42, no.~1, pp. 3--39, 2000.

\bibitem{kaufman2008staggered}
D.~M. Kaufman, S.~Sueda, D.~L. James, and D.~K. Pai, ``Staggered projections for frictional contact in multibody systems,'' in \emph{ACM SIGGRAPH Asia 2008 papers}, 2008, pp. 1--11.

\bibitem{daviet2011hybrid}
G.~Daviet, F.~Bertails-Descoubes, and L.~Boissieux, ``A hybrid iterative solver for robustly capturing coulomb friction in hair dynamics,'' in \emph{Proceedings of the 2011 SIGGRAPH Asia Conference}, 2011, pp. 1--12.

\bibitem{baraff1993issues}
D.~Baraff, ``Issues in computing contact forces for non-penetrating rigid bodies,'' \emph{Algorithmica}, vol.~10, pp. 292--352, 1993.

\bibitem{cundall1979discrete}
P.~A. Cundall and O.~D. Strack, ``A discrete numerical model for granular assemblies,'' \emph{geotechnique}, vol.~29, no.~1, pp. 47--65, 1979.

\bibitem{terzopoulos1987elastically}
D.~Terzopoulos, J.~Platt, A.~Barr, and K.~Fleischer, ``Elastically deformable models,'' in \emph{Proceedings of the 14th annual conference on Computer graphics and interactive techniques}, 1987, pp. 205--214.

\bibitem{bridson2002robust}
R.~Bridson, R.~Fedkiw, and J.~Anderson, ``Robust treatment of collisions, contact and friction for cloth animation,'' \emph{ACM Trans Graph}, vol.~21, no.~3, pp. 594--603, 2002.

\bibitem{harmon2008robust}
D.~Harmon, E.~Vouga, R.~Tamstorf, and E.~Grinspun, ``Robust treatment of simultaneous collisions,'' in \emph{ACM SIGGRAPH 2008 papers}, 2008, pp. 1--4.

\bibitem{masterjohn2022velocity}
J.~Masterjohn, D.~Guoy, J.~Shepherd, and A.~Castro, ``Velocity level approximation of pressure field contact patches,'' \emph{IEEE Robotics and Automation Letters}, vol.~7, no.~4, pp. 11\,593--11\,600, 2022.

\bibitem{muller2007position}
M.~M{\"u}ller, B.~Heidelberger, M.~Hennix, and J.~Ratcliff, ``Position based dynamics,'' \emph{Journal of Visual Communication and Image Representation}, vol.~18, no.~2, pp. 109--118, 2007.

\bibitem{macklin2016xpbd}
M.~Macklin, M.~M{\"u}ller, and N.~Chentanez, ``Xpbd: position-based simulation of compliant constrained dynamics,'' in \emph{Proceedings of the 9th International Conference on Motion in Games}, 2016, pp. 49--54.

\bibitem{bouaziz2014projective}
S.~Bouaziz, S.~Martin, T.~Liu, L.~Kavan, and M.~Pauly, ``Projective dynamics: fusing constraint projections for fast simulation,'' \emph{ACM Transactions on Graphics (TOG)}, vol.~33, no.~4, p. 154, 2014.

\bibitem{ferguson2021intersection}
Z.~Ferguson, M.~Li, T.~Schneider, F.~Gil-Ureta, T.~Langlois, C.~Jiang, D.~Zorin, D.~M. Kaufman, and D.~Panozzo, ``Intersection-free rigid body dynamics,'' \emph{ACM Transactions on Graphics}, vol.~40, no.~4, p. 183, 2021.

\bibitem{lan2022affine}
L.~Lan, D.~M. Kaufman, M.~Li, C.~Jiang, and Y.~Yang, ``Affine body dynamics: fast, stable and intersection-free simulation of stiff materials,'' \emph{ACM Transactions on Graphics (TOG)}, vol.~41, no.~4, pp. 1--14, 2022.

\bibitem{drake_fric_0}
A.~Castro, X.~Han, and J.~Masterjohn, ``A theory of irrotational contact fields,'' \emph{arXiv preprint arXiv:2312.03908}, 2023.

\bibitem{berkooz1993proper}
G.~Berkooz, P.~Holmes, and J.~L. Lumley, ``The proper orthogonal decomposition in the analysis of turbulent flows,'' \emph{Annual review of fluid mechanics}, vol.~25, no.~1, pp. 539--575, 1993.

\bibitem{barbivc2008real}
J.~Barbi{\v{c}} and J.~Popovi{\'c}, ``Real-time control of physically based simulations using gentle forces,'' \emph{ACM transactions on graphics (TOG)}, vol.~27, no.~5, pp. 1--10, 2008.

\bibitem{pentland1989good}
``Good vibrations: Modal dynamics for graphics and animation,'' in \emph{Proceedings of the 16th annual conference on Computer graphics and interactive techniques}, 1989, pp. 215--222.

\bibitem{hauser2003interactive}
K.~K. Hauser, C.~Shen, and J.~F. O’Brien, ``Interactive deformation using modal analysis with constraints,'' in \emph{Graphics Interface}, 2003, p. 247.

\bibitem{lee2020model}
K.~Lee and K.~T. Carlberg, ``Model reduction of dynamical systems on nonlinear manifolds using deep convolutional autoencoders,'' \emph{Journal of Computational Physics}, vol. 404, p. 108973, 2020.

\bibitem{chen2022crom}
P.~Y. Chen, J.~Xiang, D.~H. Cho, Y.~Chang, G.~Pershing, H.~T. Maia, M.~M. Chiaramonte, K.~Carlberg, and E.~Grinspun, ``Crom: Continuous reduced-order modeling of pdes using implicit neural representations,'' \emph{arXiv preprint arXiv:2206.02607}, 2022.

\bibitem{zong2023neural}
Z.~Zong, X.~Li, M.~Li, M.~M. Chiaramonte, W.~Matusik, E.~Grinspun, K.~Carlberg, C.~Jiang, and P.~Y. Chen, ``Neural stress fields for reduced-order elastoplasticity and fracture,'' in \emph{SIGGRAPH Asia 2023 Conference Papers}, 2023, pp. 1--11.

\bibitem{capell2002interactive}
S.~Capell, S.~Green, B.~Curless, T.~Duchamp, and Z.~Popovi{\'c}, ``Interactive skeleton-driven dynamic deformations,'' \emph{ACM transactions on graphics (TOG)}, vol.~21, no.~3, pp. 586--593, 2002.

\bibitem{narain2012adaptive}
R.~Narain, A.~Samii, and J.~F. O'brien, ``Adaptive anisotropic remeshing for cloth simulation,'' \emph{ACM transactions on graphics (TOG)}, vol.~31, no.~6, pp. 1--10, 2012.

\bibitem{li2018implicit}
J.~Li, G.~Daviet, R.~Narain, F.~Bertails-Descoubes, M.~Overby, G.~E. Brown, and L.~Boissieux, ``An implicit frictional contact solver for adaptive cloth simulation,'' \emph{ACM Transactions on Graphics (TOG)}, vol.~37, no.~4, pp. 1--15, 2018.

\bibitem{lan2021medial}
L.~Lan, Y.~Yang, D.~Kaufman, J.~Yao, M.~Li, and C.~Jiang, ``Medial ipc: accelerated incremental potential contact with medial elastics,'' \emph{ACM Transactions on Graphics}, vol.~40, no.~4, 2021.

\bibitem{tang2009c}
M.~Tang, Y.~J. Kim, and D.~Manocha, ``C 2 a: Controlled conservative advancement for continuous collision detection of polygonal models,'' in \emph{2009 IEEE International Conference on Robotics and Automation}.\hskip 1em plus 0.5em minus 0.4em\relax IEEE, 2009, pp. 849--854.

\bibitem{kuppuswamy2020soft}
N.~Kuppuswamy, A.~Alspach, A.~Uttamchandani, S.~Creasey, T.~Ikeda, and R.~Tedrake, ``Soft-bubble grippers for robust and perceptive manipulation,'' in \emph{2020 IEEE/RSJ International Conference on Intelligent Robots and Systems (IROS)}.\hskip 1em plus 0.5em minus 0.4em\relax IEEE, 2020, pp. 9917--9924.

\bibitem{drake_fric_1}
X.~Han, J.~Masterjohn, and A.~Castro, ``A convex formulation of frictional contact between rigid and deformable bodies,'' \emph{IEEE Robotics and Automation Letters}, 2023.

\bibitem{drake}
\BIBentryALTinterwordspacing
R.~Tedrake and the Drake Development~Team, ``Drake: Model-based design and verification for robotics,'' 2019. [Online]. Available: \url{https://drake.mit.edu}
\BIBentrySTDinterwordspacing

\bibitem{nvidia_physx}
\BIBentryALTinterwordspacing
{NVIDIA Corporation}, ``{NVIDIA PhysX},'' 2023, version 5.1. [Online]. Available: \url{https://developer.nvidia.com/physx-sdk}
\BIBentrySTDinterwordspacing

\bibitem{hydroelastic}
R.~Elandt, E.~Drumwright, M.~Sherman, and A.~Ruina, ``A pressure field model for fast, robust approximation of net contact force and moment between nominally rigid objects,'' in \emph{2019 IEEE/RSJ International Conference on Intelligent Robots and Systems (IROS)}.\hskip 1em plus 0.5em minus 0.4em\relax IEEE, 2019, pp. 8238--8245.

\bibitem{drake_hydroelastic}
D.~D. Team, ``Hydroelastic contact user guide,'' [Online]. Available: \url{https://drake.mit.edu/doxygen\_cxx/group\_\_hydroelastic\_\_user\_\_guide.html}.

\end{thebibliography}

\end{document}